\documentclass[10pt,twocolumn,letterpaper]{article}

\usepackage{amsmath}
\usepackage{wacv}
\usepackage{times}
\usepackage{epsfig}
\usepackage{graphicx}
\usepackage{amssymb}

\usepackage[noend]{algpseudocode}
\usepackage{algorithm}
\usepackage[noend]{algpseudocode}

\def\wacvPaperID{****} 

\wacvfinalcopy 

\ifwacvfinal
\def\assignedStartPage{1} 
\fi

\ifwacvfinal
\usepackage[breaklinks=true,bookmarks=false]{hyperref}
\else
\usepackage[pagebackref=true,breaklinks=true,colorlinks,bookmarks=false]{hyperref}
\fi

\ifwacvfinal
\else
\fi

\begin{document}

\title{Fully-simulated Integration of Scamp5d Vision System and Robot Simulator}

\author{Wen Fan\\
University of Manchester\\
School of  Electrical \& Electronic Engineering\\
{\tt\small wen.fan@postgrad.manchester.ac.uk}
\and
Yanan Liu\\
University of Bristol\\
Bristol Robotics Lab \& Visual Information Lab\\
{\tt\small yanan.liu@ieee.org}

\and
Yifan Xing\\
University of Bristol\\
Visual Information Lab\\
{\tt\small yifan.xing@bristol.ac.uk}

}

\maketitle

\begin{abstract}

This paper proposed a fully-simulated environment by integrating an on-sensor visual computing device, SCAMP, and CoppeliaSim robot simulator via interface and remote API. Within this platform, a mobile robot obstacle avoidance and target navigation with pre-set barriers is exploited with on-sensor visual computing, where images are captured in a robot simulator and processed by an on-sensor processing server after being transferred. We made our developed platform and associated algorithms for mobile robot navigation available online.
  
   

\end{abstract}

\section{Introduction}

\begin{figure}[t]
\begin{center}
    \includegraphics[width=0.8\linewidth]{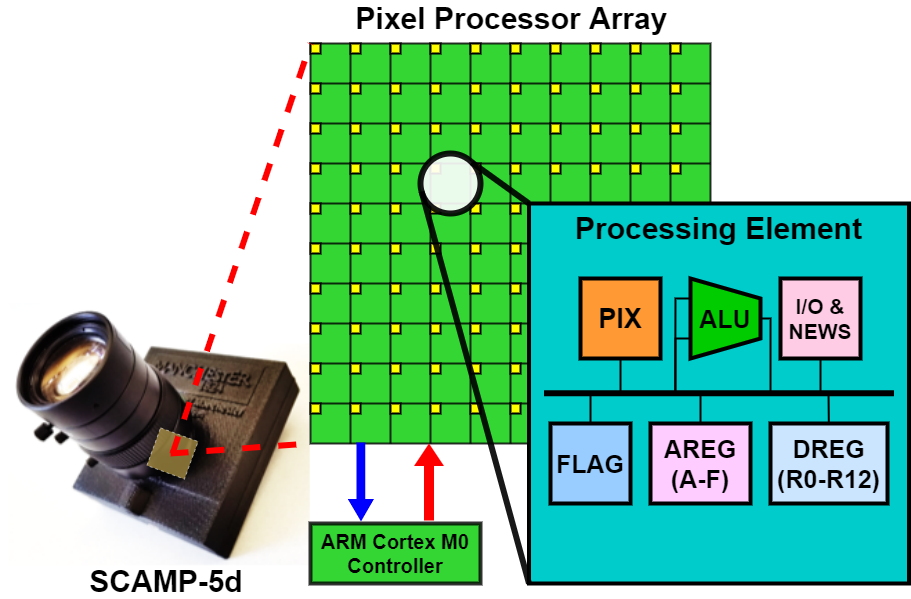}
\end{center}
   \caption{SCAMP-5d vision system and the PPA. The photosensitive chip consists of 256 $\times$ 256 processor elements where pixel, registers, arithmetic units and other peripheral units are integrated  providing parallel on-sensor computing ability. (Figure from \cite{liu2021direct})}
\label{fig:1}
\end{figure}

The Scamp vision system \cite{carey2013100,chen2018scamp5d,liu2020bmvc} (Figure~\ref{fig:1}) is a general-purpose visual device with parallel in-sensor computing capabilities enabled by novel large-scale circuit design and system integration of photosensitive pixels, registers, arithmetic units, and IO. With these features, it is increasingly integrated with robots for various applications with low-power consumption, and efficient parallel computing without external hardware. However, it is often inefficient to validate ideas and perform experiments when integrating the Scamp vision system with a mobile robot. Although an earlier semi-simulated platform is developed, the real Scamp vision system is not always accessible. Hence, authors are motivated to develop a fully-simulated platform consisting of the Scamp vision system and robot simulator for both researchers and novices to prototype ideas more efficiently and flexibly. The comparison of different platforms is shown in Table~\ref{tab:5}.

With this in mind, this work takes advantage of a comprehensive robot simulator CoppeliaSim \cite{rohmer2013v} with Scamp vision simulator, to validate ideas and develop prototype more quickly. CoppeliaSim is a virtual robot platform where each  agent  can  be  controlled via  remote  API \cite{james2019pyrep}, hence enabling sensor readings and control instructions transferred to/from other platforms. This work developed a Python-based interface \& API connecting two simulation platforms allowing data transferred mutually. In addition, through the proposed platform, we demonstrate a mobile robot navigation system based on monocular on-sensor computing with the Scamp. We made our developed platform and algorithms available from\footnote{\url{https://github.com/Neo-manchester/fully_simulated_platform_ScampSim_Coppelia.git}}. The experimental video can be seen from\footnote{\url{https://youtu.be/Ly5oBufGPag}}.

Earlier work on the platform development of Scamp vision system mainly include Chen \etal \cite{chen2018scamp5d} where the whole development framework was proposed, Liu \etal \cite{liu2021bringing} introduced a semi-simulated platform by integrating a real Scamp5d vision system with CoppeliaSim robot simulator for more user-friendly robot-related application development. Based on these two works, this paper developed a fully-simulated development platform for more flexible idea validation before mounting a real Scamp upon a mobile platform. Table~\ref{tab:5} compares these three platforms where different levels of simulation and its features are shown.



 \begin{table*}
\begin{center}
\begin{tabular}{|c|c|c|c|}
\hline
Platform & configuration & Flexibility & Developing cycle\\
\hline\hline
Real Scamp vision system \cite{chen2018scamp5d} & SCAMP + A mobile platform & Low & Long\\
Semi-sim platform \cite{liu2021bringing} & SCAMP + A robot simulator & Medium & Medium\\
Fully-sim platform (this work) & SCAMP simulator + A robot simulator & High & Short\\
\hline
\end{tabular}
\end{center}
\caption{Comparison of different simulation levels of Scamp vision system and mobile platforms.}
\label{tab:5}
\end{table*}

\section{Simulation Platform Development}

\begin{figure}[t]
\begin{center}
   \includegraphics[width=1.0\linewidth]{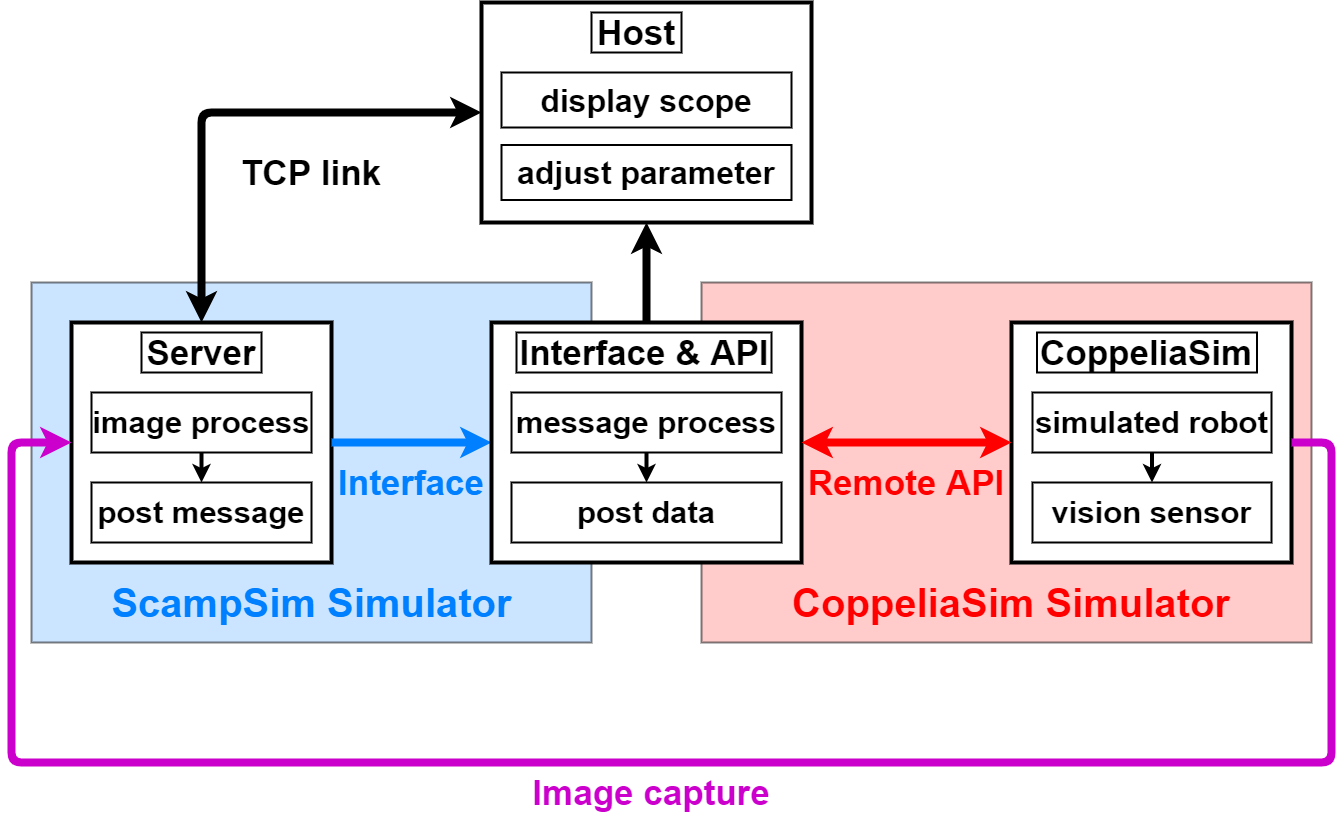}
\end{center}
   \caption{The framework of the fully-simulated system. Once the Server is started, the Host works via TCP link. Then the CoppeliaSim is switched on by the remote API. After that, a closed loop of image processing from ScampSim, 
   data transmission through interface/API, robot simulation and sensor reading from CoppeliaSim is set up, coming along with visualisation and parameter tuning on the Host.}
\label{fig:4}
\end{figure}

The framework of the fully-simulated system is shown in Figure~\ref{fig:4}, which mainly consists of ScampSim, CoppeliaSim, Interface \& API, and Host.


\subsection{SCAMP-5d Vision System Simulator}

\begin{figure}[t]
\begin{center}
   \includegraphics[width=1\linewidth]{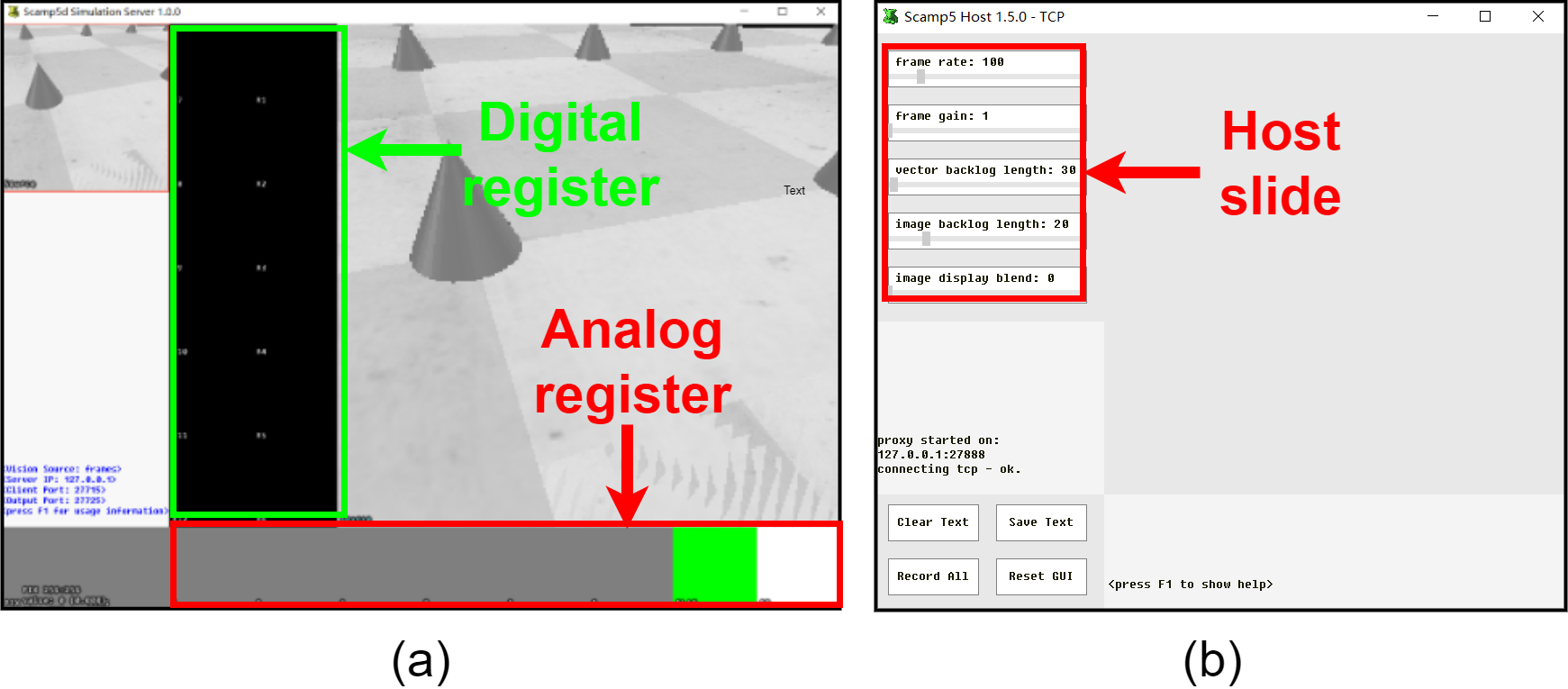}
\end{center}
   \caption{(a): ScampSim Server user interface. The image processing procedure can be seen from Digital/Analog registers. (b): The Host user interface, where customised parameters for image processing on the Server can be tuned. These two modules communicate mutually through TCP link.}
\label{fig:7}
\end{figure}

As shown in Figure~\ref{fig:4}, the ScampSim simulator usually consists of Server and communication interface with other modules. The Server can visualise the image processing and the Host provides a GUI for users to adjust image processing parameters (Figure~\ref{fig:7}). The CoppeliaSim serves as a comprehensive robot simulator where customised robot-related applications, such as vision sensors, can be set. Then, sensor readings can be transferred to the ScampSim Server for image processing. After that, the processed data are sent back through Interface \& API to the robot simulator guiding the robots performing tasks. Here, the Interface \& API acts as a 'bridge' to connect the other two modules for data transmission. The detail of the SCAMP-5d vision simulation and setup can be seen from \footnote{\url{https://scamp.gitlab.io/scamp5d_doc/_p_a_g_e__s_i_m_u_l_a_t_i_o_n.html}} and \cite{chen2018scamp5d}.


\begin{figure}[t]
\begin{center}
   \includegraphics[width=0.9\linewidth]{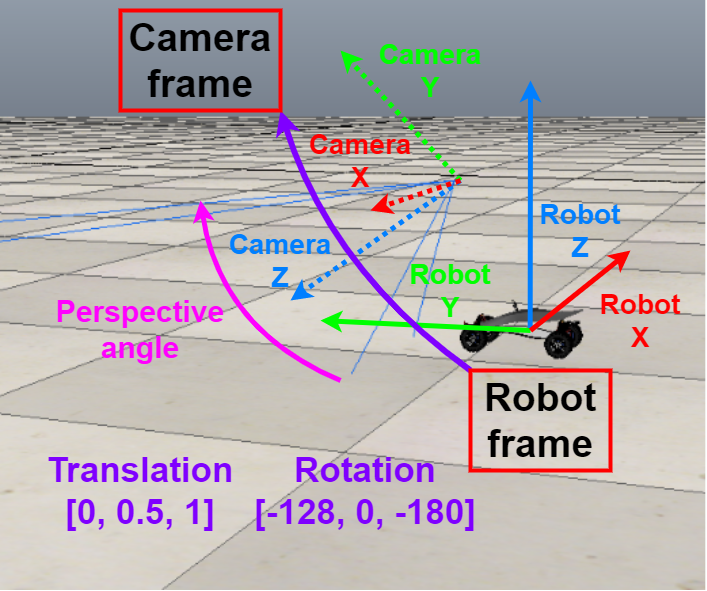}
\end{center}
   \caption{The frame difference between robot and camera in CoppeliaSim. The translation [$X$,$Y$,$Z$] and rotation [$\Psi$,$\theta$,$\Phi$] from robot to camera are defined as [0,0.5,1] and [-128,0,-180]. The camera perspective angle is 60$^\circ$ both vertically and horizontally.}
\label{fig:6}
\end{figure}

\subsection{CoppeliaSim Robot Simulation Environment}

The CoppeliaSim\footnote{\url{https://www.coppeliarobotics.com/}} \cite{rohmer2013v} is a popular robot simulator where the many types of mobile platforms and ambient environments are supported. By mounting a vision sensor and adjusting the associated parameters onto a Manta mobile robot, the camera can simulate a real Scamp5d camera for the gray-scale image of 256 $\times$ 256 capturing (Figure~\ref{fig:6}).


\subsection{The Interface \& Remote API}

The Python-based interface is introduced to receive the messages from the ScampSim Server which is similar to the earlier simulated platform \cite{liu2021bringing}. In addition, the interface integrates the remote API to communicate with the CoppeliaSim, which enabled by several CoppeliaSim API functions, such as '$client.simxGetObjectPosition$' and '$client.simxSetJointPosition$'. These functions can read and set states of agents within CoppeliaSim according to the instructions from Server through Interfaces \& API.

\begin{figure}[t]
\begin{center}
   \includegraphics[width=0.9\linewidth]{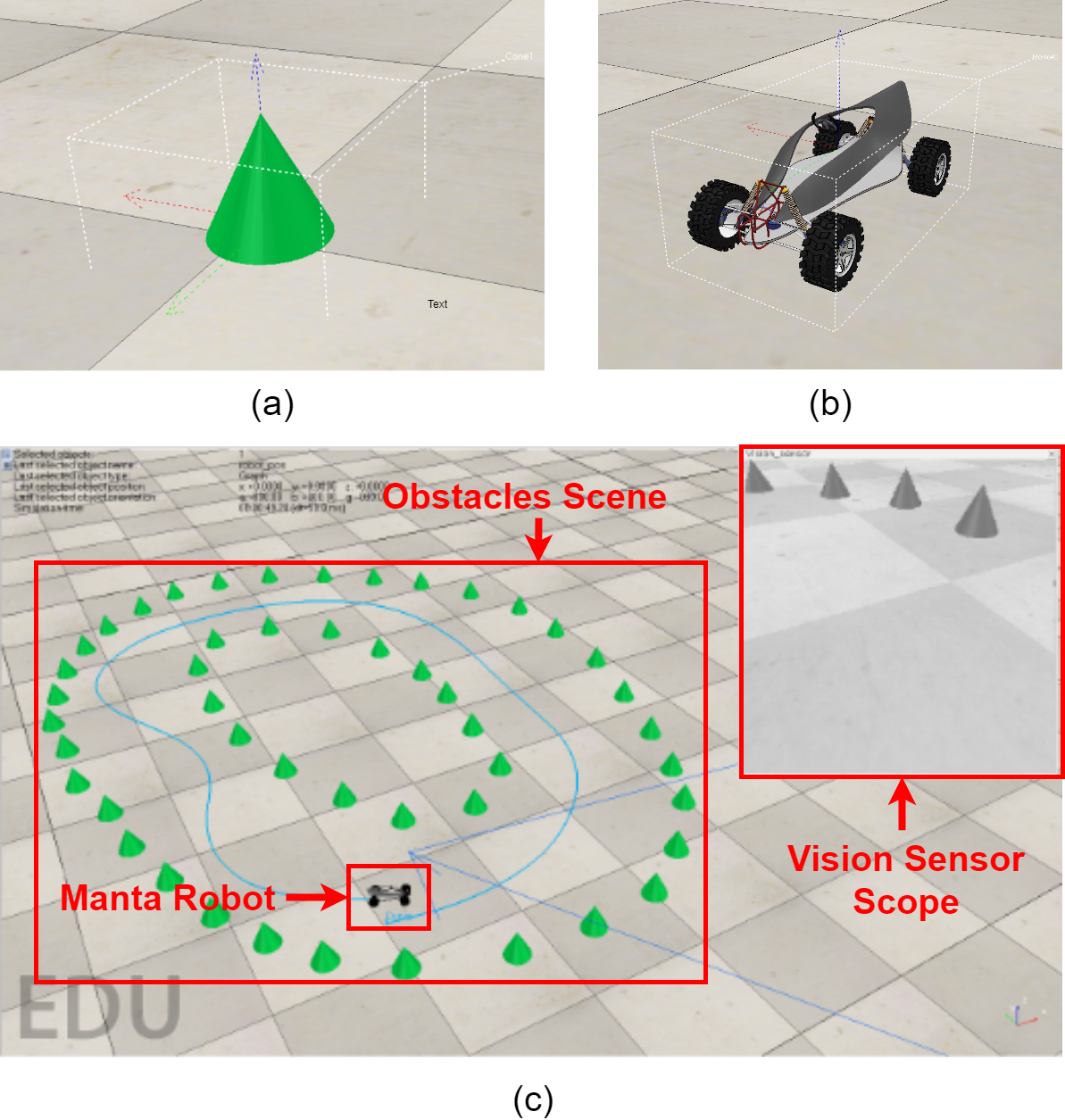}
\end{center}
   \caption{Obstacles and Robot in the CoppeliaSim. (a) Cone as the obstacles.  (b) The Manta mobile vehicle. (c) The example of experimental environment setups within the CoppeliaSim.}
\label{fig:robot_obstacle}
\end{figure}

\section{Experiment}

To validate the effectiveness of the proposed simulated platform, this paper set up a simple environment for the robot navigation testing the whole loop of the image capturing, data transmission, image processing, and robot control. As shown in Figure \ref{fig:robot_obstacle}, the cone as the obstacles can provide features such as corner points, sloped edges and round bottom. With this setup, the simulation environment can be simplified to verify the effectiveness of the system more efficiently. The Manta robot is selected in this case, a wheeled mobile robot based on the Ackerman chassis. Users can take advantage of our platform for idea validation by setting up the more complex environment.


\subsection{Robot Navigation Algorithm}

\begin{figure}[t]
\begin{center}
   \includegraphics[width=1.0\linewidth]{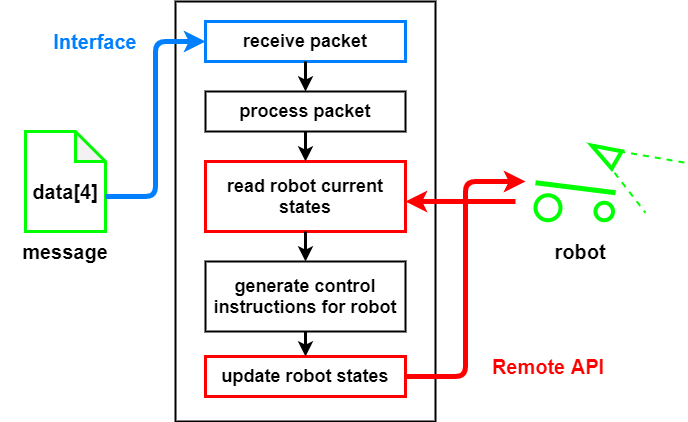}
\end{center}
   \caption{The operation process of robot navigation via interface \& API. After being received by the Interface, the message from the Server is processed to generate the control instructions for the robot in CoppeliaSim.
   }
\label{fig:8}
\end{figure}

This section proposes algorithms for the robot obstacle avoidance and navigation to the pre-set targets, which can be seen in Figure~\ref{fig:8} and Algorithm~\ref{alg:1} in detail. '$data[4]$' is the message from the Server containing robot navigation and obstacle avoidance related information:  '$closest\_x$', '$closest\_y$', '$closest\_dis$' and '$direction$' (Figure~\ref{fig:13}). $D_{safe}$ is the safe range to avoid the collision. The obstacle avoidance manoeuvre should be performed once the distance between the vehicle and barriers is detected to be smaller than the $D_{safe}$. The target is achieved successfully when the distance between the vehicle and the target is smaller than the $E_{safe}$.





\begin{algorithm}[t]
\caption{Robot Navigation Algorithm}
\label{alg:1}
\hspace*{0.02in} {\bf Input:} 
\emph{data[4]}, \emph{target}  // Server message and pre-set targets\\
\hspace*{0.02in} {\bf Output:} 
$\theta_{steer}$ // control the robot steer angle
\begin{algorithmic}[1]
\While{\emph{true}} 
    \State \emph{packet} = \emph{Scamp5\_get\_packet()} // read via interface
        \If{\emph{packet is None}} 
        \State \textbf{break}
    \Else
        \State \emph{data[4]} = \emph{process\_packet}
            \If{$D_{safe}$ \textless \, \emph{closest\_dis}} // target navigation
        \State \emph{pos} = \emph{robot coordinate} // read via API
        \State $\theta_{target}$ = \emph{atan2(target - pos)}
        \State $E_{dis}$ = \emph{sqrt(target - pos)}
            \If{$E_{safe}$ \textless \, $E_{dis}$} 
        \State $\theta_{robot}$ = \emph{robot orientation} // read via API
        \State $\theta_{steer}$ = $\theta_{target}$ - $\theta_{robot}$
        \State sent $\theta_{steer}$ to robot // write via API
    \Else
        \State \textbf{break} // on target
    \EndIf
    \Else \, // obstacle avoidance navigation
        \State $\theta_{steer}$ = \emph{control output (Algorithm 2)}
        \State sent $\theta_{steer}$ to robot // write via API
    \EndIf
    \EndIf
\EndWhile
\State \Return $\theta_{steer}$

\end{algorithmic}
\end{algorithm}


\subsection{Image Processing Algorithm}

\begin{figure}[t]
\begin{center}
   \includegraphics[width=1.0\linewidth]{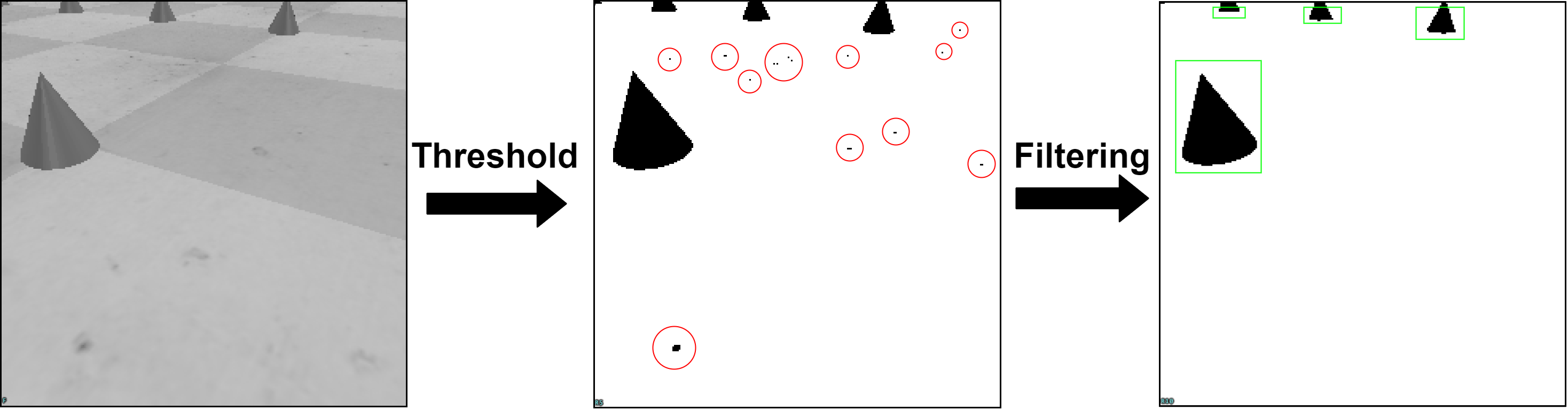}
\end{center}
   \caption{Threshold and noise filtering for obstacle segmentation.}
   
\label{fig:10}
\end{figure}

The ScampSim Server performs image processing algorithms in the same parallel way with the real Scamp vision system once the sensor reading image is received from the CoppeliaSim. As can be seen from Figure \ref{fig:10}, binary thresholding and noise filtering are the main image preprocess methods to efficiently segment obstacles out with a clean background for further image processing.



\begin{figure}[t]
\begin{center}
   \includegraphics[width=1.0\linewidth]{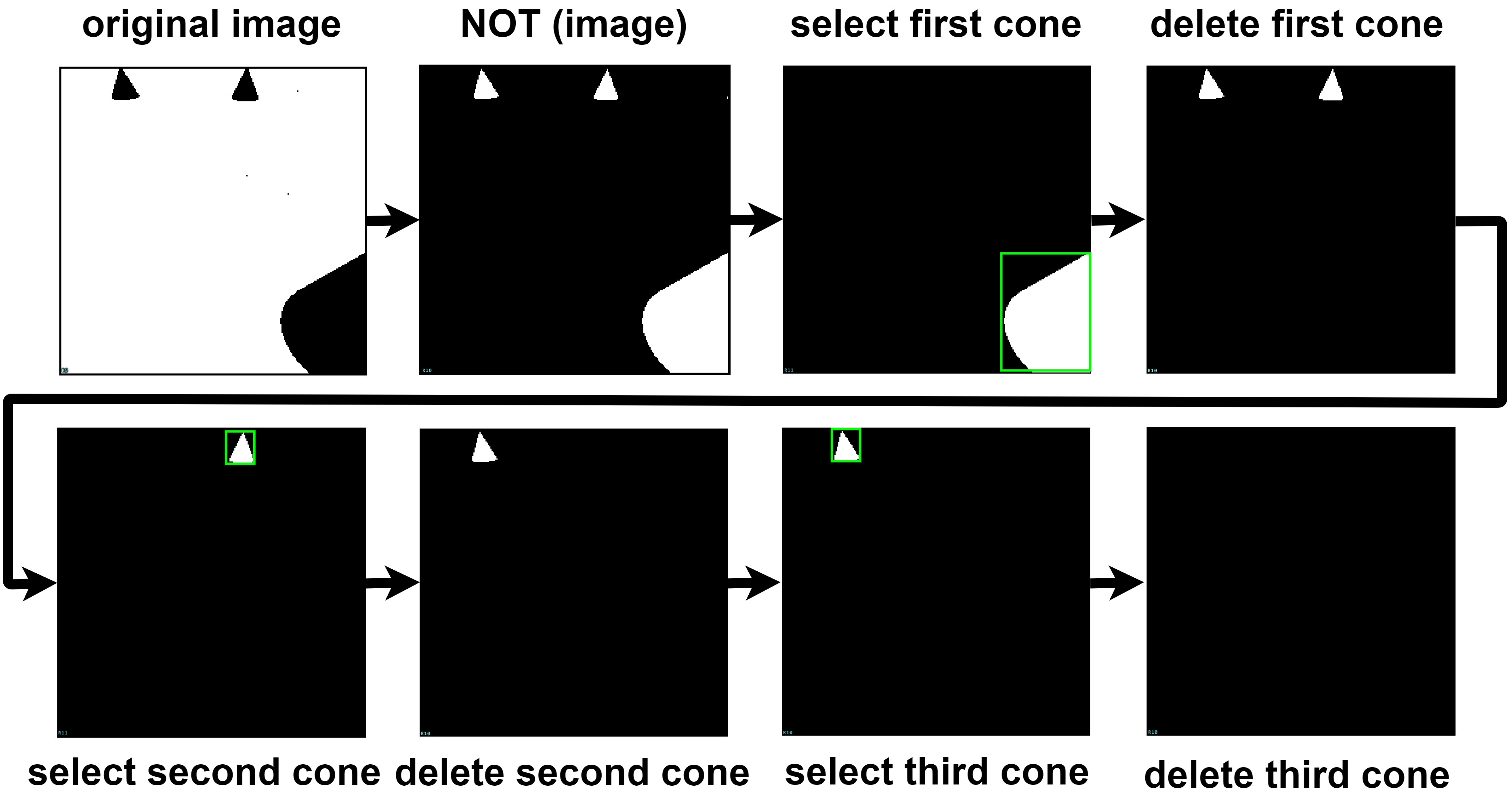}
\end{center}
   \caption{The operation process of cone boundingbox extraction.}
\label{fig:12}
\end{figure}
\begin{figure}[t]
\begin{center}
   \includegraphics[width=1.0\linewidth]{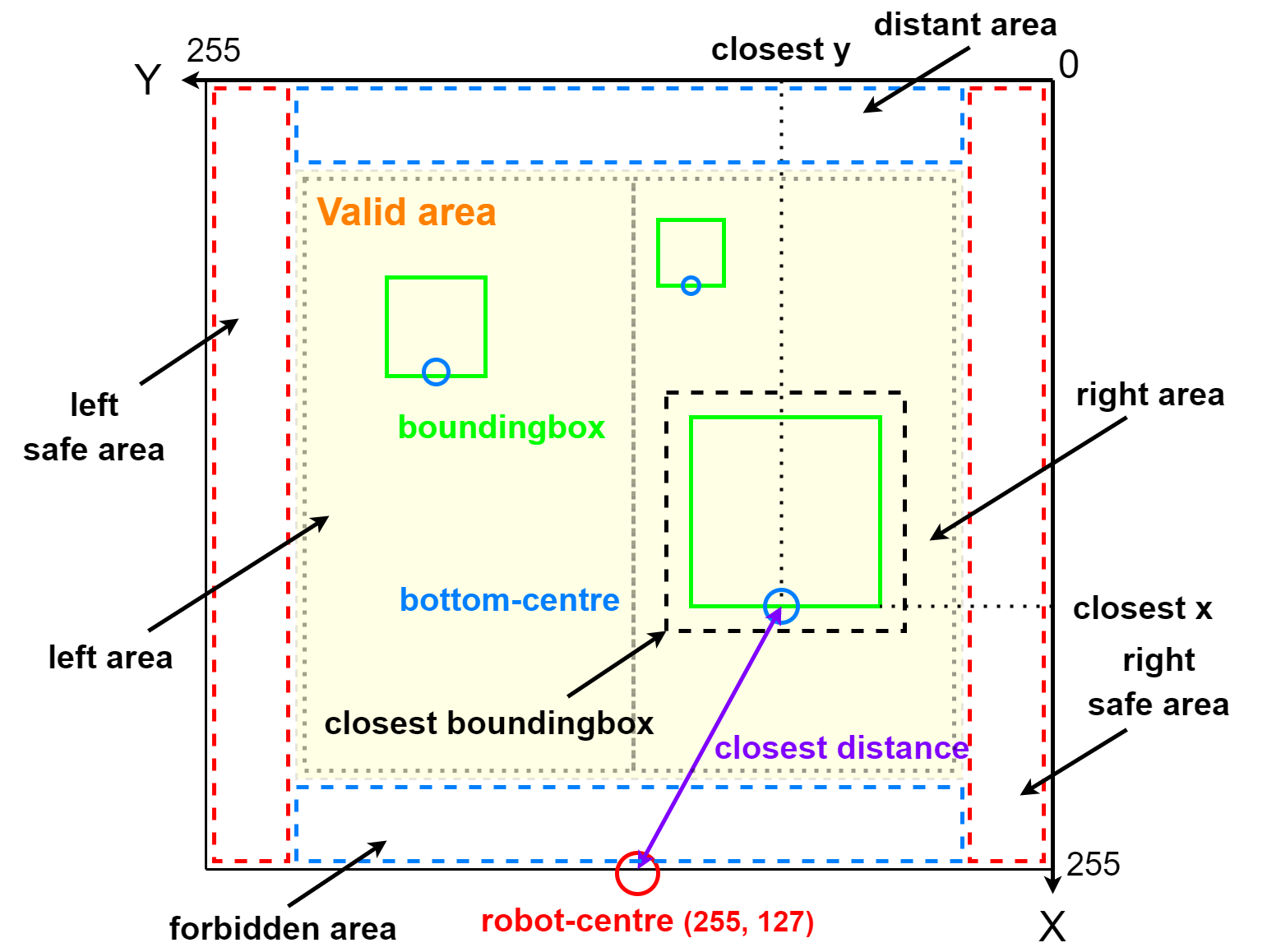}
\end{center}
   \caption{The principle of the closest obstacle detection algorithm. The key idea is to find out the closest boundingbox to (255, 127), which indicates the robot position. The closest distance will be updated if the bottom centre exists inside the valid area, except for ‘safe area’, ‘forbidden area’ and ‘distant area’. The right and left areas present the right and left parts of the valid area.}
\label{fig:13}
\end{figure}

Then, this work takes advantage of the Scamp built-in flooding function ($scamp5\_flood$) to efficiently obtain the bounding box ('$scamp5\_scan\_boundingbox$') of each cone, hence to get the approximate size information of each obstacle with the thinking of a bigger bounding box indicating a shorter distance. The obstacle avoidance starts from the closest one. Details can be seen from Algorithm~\ref{alg:4}, Figure~\ref{fig:12} and Figure~\ref{fig:13}.


Once cones are segmented, the first '1' pixel can be the source to apply the flooding operation to get rid of other cones. The boundingbox function is performed to get the size of this cone (Figure~\ref{fig:12}). Then, the found cone should be deleted from the original image to start the next iteration of cone detection until all the cones are extracted.


\begin{algorithm}[t]
\caption{Closest obstacle detection algorithm} 
\label{alg:4}
\hspace*{0.02in} {\bf Input:} 
\emph{R10} // image after threshold and filtering\\
\hspace*{0.02in} {\bf Output:} 
\emph{data[4]} // post message
\begin{algorithmic}[1]
\While{\emph{OR(NOT(R10)) = true}} 
    \State \emph{event\_coords[count]} = \emph{Scamp5\_scan\_evens(R10,1)} 
    \State \emph{R11} = \emph{Scamp5\_load\_point(event\_coords[count])} 
    \State \emph{R11} = \emph{Scamp5\_flood(R10)} 
       \State \emph{boundingbox} = \emph{Scamp5\_scan\_boundingbox(R11)}

                    \If{\emph{distant area} \textless \, $X_{bottom}$ \textless \, \emph{forbidden area}} 
                    \If{\emph{right safe} \textless \, $Y_{bottom}$ \textless \, \emph{left safe}} 
        \State \emph{dis} = \emph{sqrt((255,127) - ($X_{bottom}$,$Y_{bottom}$))}
            \If{\emph{dis} \textless \, \emph{closest\_dis}} 
        \State \emph{closest\_dis} = \emph{dis}
        \State \emph{closest\_x} = \emph{$X_{bottom}$}
        \State \emph{closest\_y} = \emph{$Y_{bottom}$}
    \Else
    \EndIf
    \Else 
    \EndIf
    \Else

    \EndIf
        \State \emph{direction} = \emph{closest\_y whether in right or left area}
        \State \emph{R10} = \emph{XOR(R10,R11)}
        \State \emph{count++}
        \State \emph{data[4]} = \emph{[closest\_x,closest\_y,closest\_dis,direction]}
        \State \emph{post(data[4])} // post message to interface 
\EndWhile
\State \Return \emph{data[4]}

\end{algorithmic}
\end{algorithm}

\subsection{Robot Obstacle Avoidance Algorithm}

If no barriers are detected, the robot will follow the target navigation algorithm shown in Algorithm~\ref{alg:1}. Once obstacle distances are smaller than $D_{safe}$, the obstacle avoidance functions are activated shown in Equation \ref{1} and \ref{2}. 

\begin{equation}
\mathit{K} = O_{closest} \times \left ( \frac{K_{safe} }{D_{closest}}  \right ) \label{1}
\end{equation}

\begin{equation}
\theta _{steer} = - K_{steer} \times K  \label{2}
\end{equation}
The $D_{closest}$ and $O_{closest}$ represent the closest obstacle’s distance and direction in message \emph{data[4]}. A closer cone produces a bigger $\theta_{steer}$. The $K_{safe}$  and $K_{steer}$ are adjustable parameters related to the safe range length $D_{safe}$ and steering response sensitivity respectively. The negative sign in Equation \ref{2} means the robot turns towards the direction far from the closest obstacle. 


\subsection{Navigation Experiments}

\begin{figure}[t]
\begin{center}
   \includegraphics[width=1.0\linewidth]{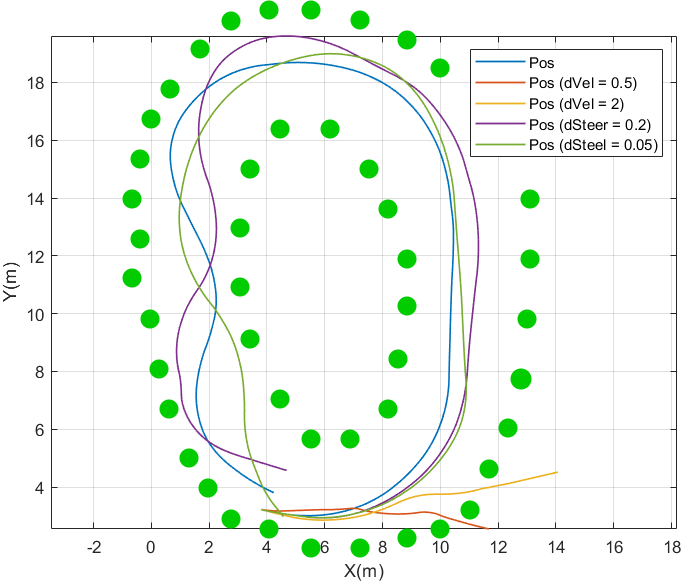}
\end{center}
   \caption{The robot trajectories in the reactive navigation. 
   }
\label{fig:16}
\end{figure}

\begin{figure}[t]
\begin{center}
   \includegraphics[width=1.0\linewidth]{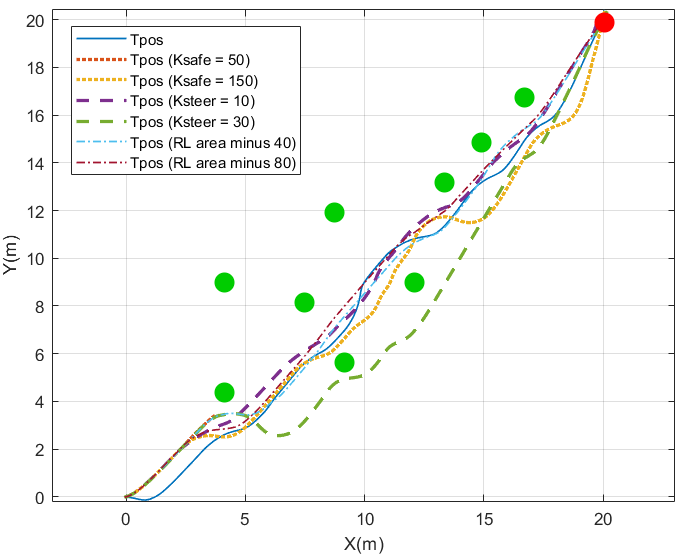}
\end{center}
   \caption{The robot trajectories in the single target navigation. 
   }
\label{fig:19}
\end{figure}

\begin{figure}[t]
\begin{center}
   \includegraphics[width=1.0\linewidth]{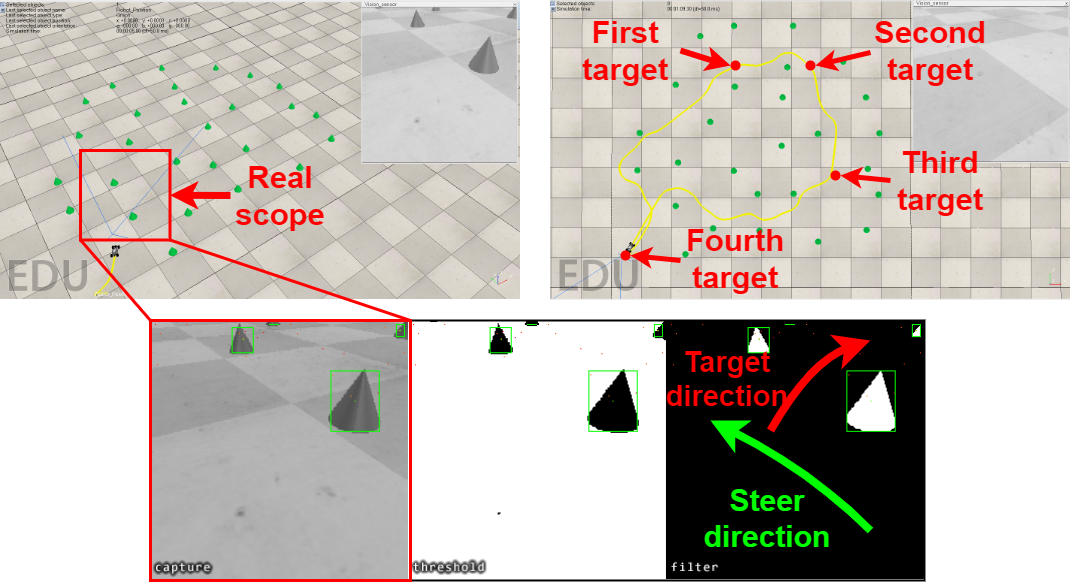}
\end{center}
   \caption{The simulation of the multiple target navigation. 
   }
\label{fig:22}
\end{figure}

\begin{figure}[t]
\begin{center}
   \includegraphics[width=0.9\linewidth]{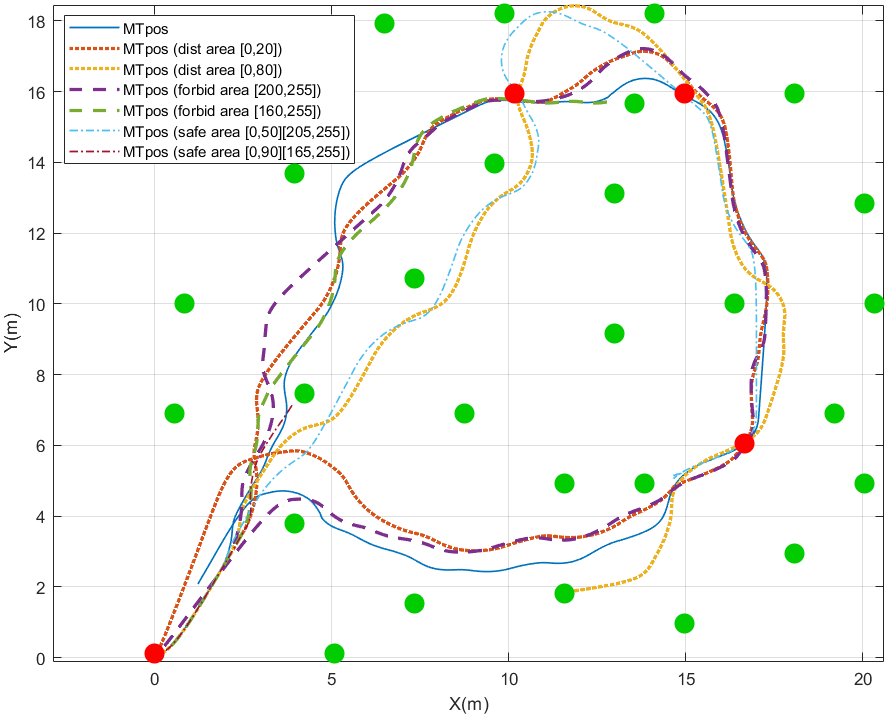}
\end{center}
   \caption{The robot trajectories in the multiple target navigation. 
   }
\label{fig:23}
\end{figure}

This section presents three navigation tasks, reactive navigation \cite{liu2021agile}, single target, and multiple targets navigation, to verify the fully-simulated environment and the above-mentioned algorithms. The initial parameters for the experiment setup are shown in Table~\ref{tab:3}. For the reactive navigation, the robot navigates freely to prevent collisions in the environment with an obstacle layout of concentric-like ellipses as shown in Figure \ref{fig:16} where the multiple trajectories can be seen. Then single and multiple tasks navigation are illustrated. In this case, the robot not only performs obstacle avoidance but also path planning towards its target position through obstacles (Figure~\ref{fig:19},Figure~\ref{fig:22} and Figure~\ref{fig:23}).

\begin{table}
\begin{center}
\begin{tabular}{|l|c|}
\hline
Parameter ID & Initial Value \\
\hline\hline
dVel & 1(r/s) \\
dSteer & 0.1(rad) \\
$K_{safe}$ & 100\\
$K_{steer}$ & 20\\
$D_{safe}$ & 200\\
$E_{safe}$ & 1\\
distant area & [0,50]\\
forbidden area & [240,255]\\
safe area & [0,10],[245,255]\\
right and left area & [15,127],[128,240]\\
\hline
\end{tabular}
\end{center}
\caption{The initial parameters of the fully-simulated system}
\label{tab:3}
\end{table}





\section{Conclusion and Future Work}

A fully-simulated environment integrating a Scamp simulator and CoppeliaSim is presented in this work providing researchers with a flexible idea and prototype validation platform. In the experiment, we use reactive navigation, single task and multi-task navigation to validate the effectiveness of the proposed platform. Customised environments and mobile platforms can be built based on our proposed platform by users according to different types of applications.




However, a more effective synchronisation mechanism should be optimised to decrease the influence of communication latency among these three modules. In addition, the Interface \& remote API module can be improved along with the updates of Interface libraries for the Scamp vision system in the future.




{\small
\bibliographystyle{ieee_fullname}
\bibliography{egbib}
}


\end{document}